\documentclass{bmvc2k}
\usepackage{pifont}
\usepackage{float}
\usepackage{xcolor}
\usepackage{bm}
\usepackage{multicol}
\usepackage{multirow}
\usepackage{array}
\usepackage{url}
\usepackage{booktabs}
\usepackage{soul}
\usepackage[inline]{enumitem}
\usepackage{arydshln}
\usepackage[ruled,vlined]{algorithm2e}

\SetCommentSty{mycommfont}

\newcommand\blfootnote[1]{%
  \begingroup
  \renewcommand\thefootnote{}\footnote{#1}%
  \addtocounter{footnote}{-1}%
  \endgroup
}

\title{AEI: Actors-Environment Interaction with Adaptive Attention for Temporal Action Proposals Generation}

\addauthor{Khoa Vo}{khoavoho@uark.edu}{1}
\addauthor{Hyekang Joo$^\ast$}{hkjoo@cs.umd.edu}{2}
\addauthor{Kashu Yamazaki$^\ast$}{kyamazak@uark.edu}{1}
\addauthor{Sang Truong}{sangt@uark.edu}{1}
\addauthor{Kris Kitani}{kkitani@cs.cmu.edu}{3}
\addauthor{Minh-Triet Tran}{tmtriet@hcmus.edu.vn}{4,5}
\addauthor{Ngan Le}{thile@uark.edu}{1}

\addinstitution{
 AICV Lab, University of Arkansas\\
 Fayetteville, AR, USA
}
\addinstitution{
 University of Maryland\\
 College Park, MD, USA
}
\addinstitution{
 Carnegie Mellon University\\
 Pittsburgh, PA, USA
}
\addinstitution{
 University of Science VNU-HCM\\
 Ho Chi Minh City, Vietnam
}
\addinstitution{
 Vietnam National University\\
 Ho Chi Minh City, Vietnam
}

\runninghead{Khoa Vo et al.}{AEI with Adaptive Attention}

\begin{document}

\maketitle

\begin{abstract}
Humans typically perceive the establishment of an action in a video through the interaction between an actor and the surrounding environment. An action only starts when the main actor in the video begins to interact with the environment, while it ends when the main actor stops the interaction. Despite the great progress in temporal action proposal generation, most existing works ignore the aforementioned fact and leave their model learning to propose actions as a black-box. In this paper, we make an attempt to simulate that ability of a human by proposing Actor Environment Interaction (\textbf{AEI}) network to improve the video representation for temporal action proposals generation. AEI contains two modules, i.e., perception-based visual representation (PVR) and boundary-matching module (BMM). PVR represents each video snippet by taking human-human relations and humans-environment relations into consideration using the proposed \textbf{adaptive attention mechanism}. Then, the video representation is taken by BMM to generate action proposals. AEI is comprehensively evaluated in ActivityNet-1.3 and THUMOS-14 datasets, on temporal action proposal and detection tasks, with two boundary-matching architectures (i.e., CNN-based and GCN-based) and two classifiers (i.e., Unet and P-GCN). Our AEI robustly outperforms the state-of-the-art methods with remarkable performance and generalization for both temporal action proposal generation and temporal action detection. Source code is available at \footnote{\url{https://github.com/vhvkhoa/TAPG-AgentEnvInteration.git}}.
\blfootnote{$^\ast$These authors contributed equally.}
\end{abstract}

%-------------------------------------------------------------------------

\vspace*{-\baselineskip}

\begin{figure}[t]
\centering
  \includegraphics[width=0.7\linewidth]{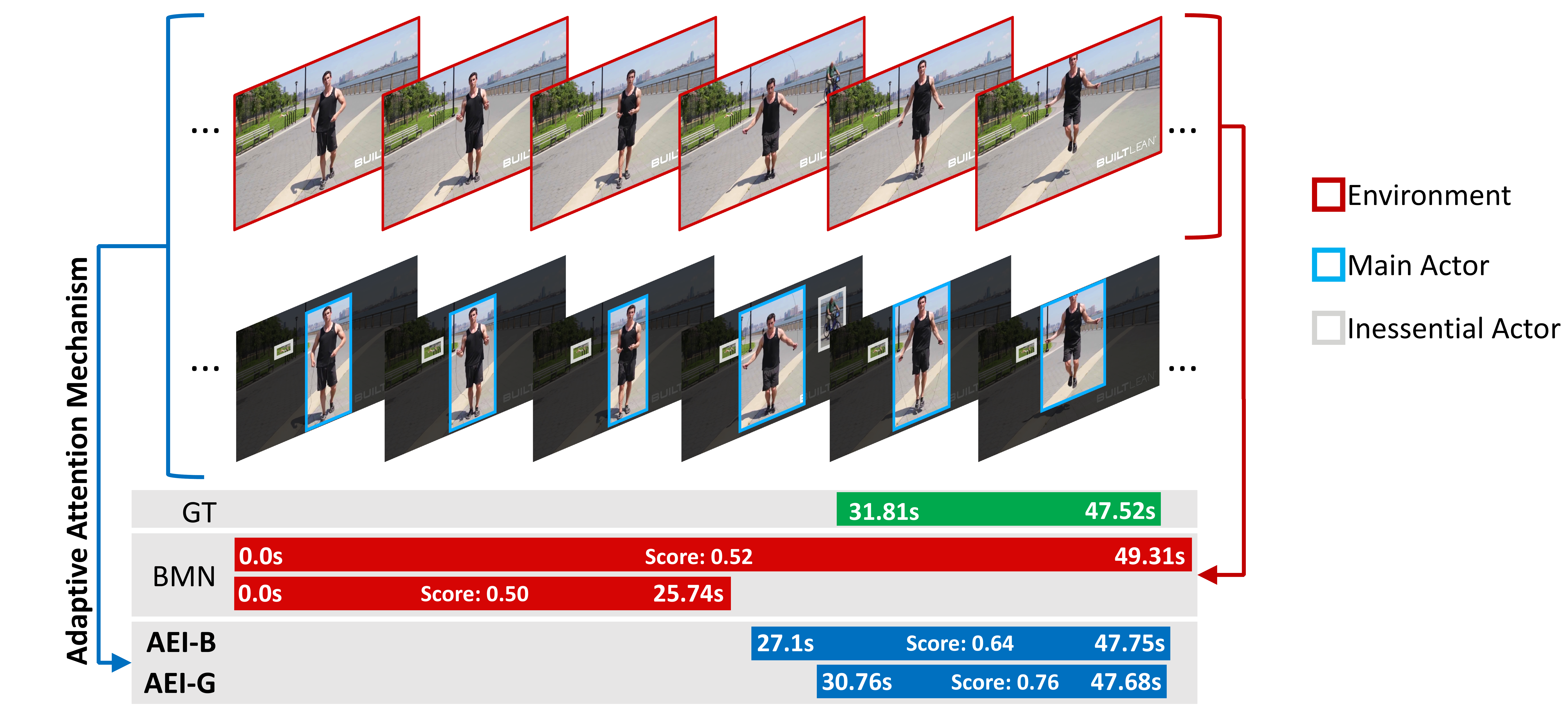}
  \vspace*{-0.1cm}
  \caption{TAPG comparison. The existing approaches (e.g., BMN \cite{bmn}) apply a network backbone to the entire spatial domain (red boxes); Our AEI takes main actor(s) into consideration via AAM. Our AEI is implemented with CNN-based BMM (AEI-B) and GCN-based BMM (AEI-G).}
  \vspace*{-0.4cm}
  \label{qualitative}
\end{figure}

%\vspace*{-\baselineskip}

%%%%%%%%% BODY TEXT
\section{Introduction}
\label{sec:intro}
%Temporal action proposals generation (TAPG) \cite{lin2018bsn, dbg, BSN++, CTAP, SRG, actionproposal_2016, bmn} is one of the most important and fundamental problems in video analysis and video understanding \cite{anchor_2, Jiyang2017, CTAP, Gao_2018_CVPR}. Given an untrimmed video, TAPG aims to propose temporal intervals with specific starting and ending timestamps for each action. Most of existing TAPG works first detect a set of possible starting and ending timestamps of all actions separately, and then a proposal evaluation module is employed to evaluate every possible pair of starting and ending timestamps by predicting its confidence score. The non-maximum suppression (NMS) function is finally used to eliminate redundant candidate proposals based on their confidence scores and overlapping metrics.

Temporal action proposals generation (TAPG) is one of the most important problems in video analysis and video understanding \cite{anchor_2, Jiyang2017, CTAP, Gao_2018_CVPR}. Particularly, TAPG is the fundamental step for other downstream tasks, including temporal action detection \cite{caba2015activitynet, THUMOS14}, action recognition \cite{Kinetics}, and video dense captioning \cite{krishna2017dense}. Given an untrimmed video, TAPG aims to propose temporal segments with specific starting and ending timestamps for each action of interest appearing in the video.

Recently, state-of-the-art (SOTA) methods \cite{lin2018bsn, bmn, BSN++, dbg} follow a paradigm where a set of possible starting and ending timestamps of all actions are detected separately, then a proposal evaluation module is employed to evaluate every possible pair of starting and ending timestamps by predicting its confidence score. A non-maximum suppression (NMS) \cite{NMS, SoftNMS} function is finally used to eliminate redundant candidate proposals based on their confidence scores and overlapping metrics.

As we observe, a human has a capacity to perceive an action being established in a video \cite{cognitive2, cognitive_vision} in two steps. First, the main actors at each temporal period are identified; then, the interactions between main actors and the environment are observed to specify when the action starts and ends. Despite good achievements on benchmarking datasets \cite{caba2015activitynet, THUMOS14}, the SOTA approaches \cite{bmn, BSN++, dbg} disregard the above perception process of humans by only applying a backbone network (pre-trained on action recognition task) to extract the video representation, leading to a potential loss of some proposals. For instance, in Fig.~\ref{qualitative}, the works in the literature take the whole spatial region of video frames (e.g., red boxes) to propose action intervals; this, however, may lead to inaccurate results because the background occupies much bigger region than the actor who performs the action (e.g., blue boxes). In Fig.~\ref{qualitative}, the "rope skipping" action can trick an action proposal model into missing the time at which this action starts or ends due to a subtle difference in shape between between "jumping" and "standing".

In this paper, we propose a novel \textbf{Actor Environment Interaction network (AEI)} in an attempt to simulate and explore the capability of human-perception process. Our AEI consists of a visual representation module (PVR) and a boundary-matching module (BMM). The PVR is comprised of three components: (i) environment spectator; (ii) actors spectator; and (iii) actors-environment interaction spectator. As illustrated in Fig.~\ref{qualitative}, the environment spectator processes the entire spatial dimensions of a snippet (red boxes) to capture the global environmental information. The actors spectator focuses on actors (blue and light gray boxes around humans) to capture local appearance and motion information of actors. Additionally, we introduce a novel \textbf{adaptive attention mechanism (AAM)} in the actors spectator to select the main actors (blue boxes) who mainly commit the action. Given a video snippet, the features corresponding to the global environment and the local main actors are first extracted by the first two spectators. The relationship between the environment and the main actors is then modeled by the third component (i.e., the actors-environment interaction spectator).

\textbf{Our contributions are summarized as follows:}
\begin{itemize}[leftmargin=*,noitemsep,topsep=-3pt] %[leftmargin=*]
    \item We propose a video representation network, AEI, which follows the human-perception process to understand human action.
    \item We introduce a novel \emph{adaptive attention mechanism (AAM)} that simultaneously selects main actors and eliminates inessential actor(s) and then extracts semantic relations between main actors.
    \item We investigate the effectiveness of the proposed AEI by implementing the BMM under two network architectures: CNN-based and GCN-based.
    %\item We study the success of the proposed adaptive attention mechanism by comparing it with hard attention and soft attention mechanisms.
    \item Our proposed AEI network achieves the SOTA performance on common benchmarking datasets of ActivityNet-1.3 and THUMOS-14 in both TAPG and TAD tracks with a large margin compared to the previous works.
\end{itemize}

%-------------------------------------------------------------------------
\section{Related Works}

\noindent
\textbf{Temporal Action Proposal Generation (TAPG)} 

TAPG aims to propose intervals that tightly contain actions in a long untrimmed video. Previous works can be divided into two main groups: anchor-based and boundary-based. Anchor-based methods \cite{actionproposal_2016, FasterR_CNN_Action, anchor_1, anchor_2, anchor_3}, which are inspired by anchor-based object detection methods in 2D images \cite{FasterRCNN, RetinaNet, yolov3}, predefine a set of fixed segments and try to fit them into ground-truth action segments in the video. Although a regression network may be applied in some of those methods to refine the proposals, a finite number of anchors cannot fit all ground-truth actions with diverse lengths. Boundary-based methods \cite{lin2018bsn, BSN++, bmn, dbg, xu2020gtad, KhoaVo_ICASSP, KhoaVo_Access} address this problem by localizing the starting and ending timestamps of all actions appearing in the video and matching them by a boundary-matching module. Our boundary-matching module belongs to the second group.

\vspace{0.2cm}
\noindent
\textbf{Attention Networks}

Attention Networks (AN) have a long history in the artificial neural networks literature~\cite{itti1998model}. We can divide AN into two main groups: Soft-Attention Networks (Soft-AN) and Hard-Attention Networks (Hard-AN). \cite{bahdanau2014neural} was one of the first Soft-AN that was applied to machine translation. Because of its differentiable architecture, which helps the whole model learn in an end-to-end fashion, Soft-AN has become an essential component in a large number of applications (e.g., speech \cite{cho2015describing}, NLP \cite{galassi2020attention}, computer vision \cite{chaudhari2019attentive}).
%Notably, \cite{attention_is_all_you_need} proposes a self-attention network based on Soft-AN in machine translation. Because of its flexible capability to model the relations between elements of input, \cite{attention_is_all_you_need} has been popularly applied into language models and computer vision.
% Because self-attention networks are able to learn the relations between input elements regardless of their amount, its popularity is increasing in not only language models but also in computer vision.
Hard-AN was first introduced in \cite{xu2015show} and \cite{Saccader_NIPS2019} for digit and object classifications, respectively. %In these works, a policy network is trained by reinforcement learning methods \cite{} to recurrently attend to different regions of the image for several timesteps before classifying it. 
Hard-AN aims to mask out irrelevant elements of the inputs to reduce the distractions. This is an advanced benefit over Soft-AN; however, Hard-AN in \cite{xu2015show} is indifferentiable. Recently, \cite{patro2018differential} proposes a Hard-AN that can be trained by normal gradient back-propagation, with a fundamental observation that the L2-norm values of more important features are usually higher than those of less important features in a feature map.
%Most recently, \cite{KhoaVo_Access, KhoaVo_ICASSP} employs Soft-AN to balance information between environment and actors in the videos. However, these methods cannot remove noise information created by insignificant actors.
In this work, we propose AAM, a module to leverage both the differentiable Hard-AN \cite{patro2018differential} and the self-attention network \cite{attention_is_all_you_need} to select the main actors of the video and to learn the relations between main actors, respectively.
%, which is beneficial for downward modules to propose actions. Detailed information is discussed is in Sec. \ref{subsubsec:actors}.

% \hl{Add an example to visually distinguish between soft and hard}

% are initially introduced in \cite{} and \cite{} for digit classification and object classification tasks, respectively. In these works, a policy network is trained by reinforcement learning methods \cite{} to recurrently attend to different regions of the image for several timesteps before classifying it. Because hard-attention networks \cite{} are indifferentiable, they are very hard to be employed in other problems that require a large and deep network. Therefore, \cite{} proposes a hard-attention network that can be trained by normal gradient back-propagation method with an observation of the L2-norm values of more important features in a feature map are usually higher than less important features. In this work, we adapt the differentiable hard-attention proposed in \cite{} our hard attention module to select main actors in the video, which is very crucial for both temporal action proposal and detection.

\section{Our Proposed AEI}
%\subsection{Problem Statement}
%\label{subsec:problem}

Given an input video $\mathcal{V}=\{v_i\}_{i=1}^{N}$, where $N$ is the number of frames, we follow the common paradigm from previous works to divide it into a sequence of \textit{snippets}, each of which consists of $\delta$ consecutive frames from the video, resulting in a total of $T=\bigr\lceil \frac{N}{\delta} \bigr\rceil$ snippets. Let $\phi(.)$ be an encode function to extract visual representation of a $\delta$-frame snippet $s_i$, the entire video is presented as:
\begin{equation}
        f_i  =\phi(s_i), \text{and } \mathcal{F} =\{f_i\}_{i=1}^{T}
\end{equation}

%Given an input video $\mathcal{V}=\{v_i\}_{i=1}^{N}$, where $N$ is the number of frames, we follow a common paradigm from previous works to divide it into a sequence of \textit{snippets}. Each snippet $s_j$ consists of $\delta$-frames and is presented as $s_j = \{v_i|v_i \in \mathcal{V}, (j-1)\delta \leq i < j\delta\}$
%, where $j\in[1, T]$ and $T=\bigr\lceil \frac{N}{\delta} \bigr\rceil$. Denote $\mathcal{S}=\{s_j\}_{j=1}^{T}$ as a sequence of snippets obtained from the video. Let $\phi(.)$ be an encode function aiming to extract visual representation of a $\delta$-frame snippet $s_j$. Thus, visual features of a snippet $s_j$ and the entire video are presented as:
%\begin{equation}
%        f_i  =\phi(s_i), \text{and } \mathcal{F} =\{f_i\}_{i=1}^{T}
%\end{equation}

Prior works \cite{bmn, lin2018bsn, xu2020gtad} employ a pre-trained backbone network (e.g., C3D network \cite{C3D} or Two-Stream network \cite{2_stream_1}) to model $\phi(.)$.
However, simply applying those networks for video representation may have some drawbacks as mentioned in Section \ref{sec:intro}. In Section \ref{subsec:pvr}, our proposed perception-based visual representation (PVR) is discussed as an alternative to the former strategy. Then, boundary-matching module for temporal action proposals generation is discussed in Section \ref{subsec:bmm}. %; and (iii) temporal action detection module. Each component will be explained in detail below.

\begin{figure}[t]
\centering
  \includegraphics[width=0.8\linewidth]{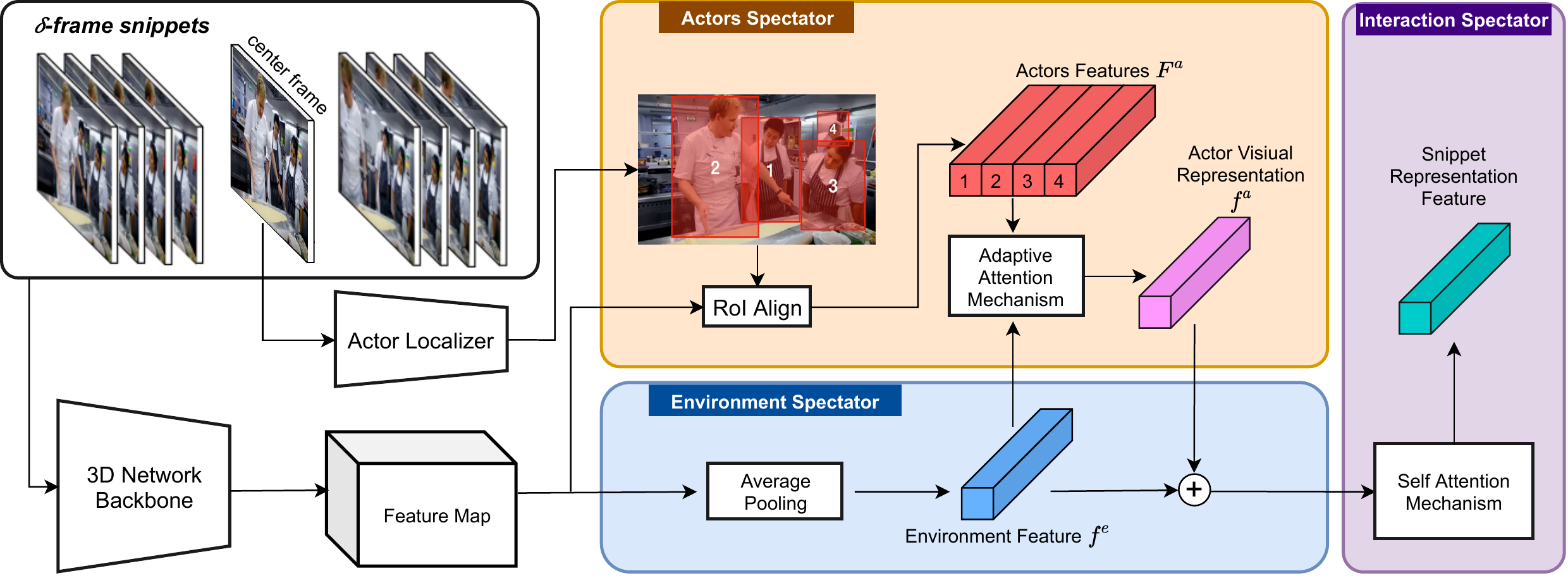}
  \vspace*{0.2cm}
  \caption{The architecture of our proposed PVR. Given a $\delta$-snippet, the corresponding snippet visual representation is obtained by three modules: (i) environment spectator to extract global environment feature; (ii) actors spectator to extract local actor representation; and (iii) actors-environment interaction spectator to model the relationship between the environment feature and the actor feature.}
  \vspace*{-0.4cm}
  \label{video_representation}
\end{figure}

\subsection{Perception-based Visual Representation (PVR)}
\label{subsec:pvr}
The PVR module aims to extract video visual representation based on how a human perceives an action, i.e., identifying the main actors at each temporal period and interactions between main actors and the environment to specify when the action starts and ends. PVR consists of three main components: (i) environment spectator; (ii) actors spectator; and (iii) actors-environment interaction spectator. The overall architecture of PVR is shown in Fig. \ref{video_representation}.

%As illustrated in Fig. \ref{video_representation}, the input $\delta$-frame snippet i(e.g., Environment Spectator, Actors Spectator, and Actors-Environment Interaction Spectator), each of which processes the input video at a certain level to comprehend relevant information that is crucial for later utilization. The first and second agents are responsible for capturing information from the whole spatial scene and individual actors, respectively. Afterwards, the third agent is accountable for combining information extracted by the first two agents.

%\subsubsection{Environment Spectator}
\vspace{0.2cm}
\textbf{(i) Environment Spectator}
\label{subsubsec:env}
aims to extract global semantic information of the input $\delta$-frame snippet. To extract both the spatial and temporal details of the snippet, we adopt a 3D network pre-trained on action recognition benchmarking datasets as a backbone feature extractor. The snippet is processed through all convolutional blocks of the 3D network to obtain a feature map $\mathcal{M}$; then, an average pooling operator is employed to produce a spatio-temporal feature vector $f^{e}$.

%To encode both the appearance and motion taking place in a $\delta$-frame snippet, we adapt a 3D network pre-trained on action recognition benchmarking datasets as a backbone feature extractor. Since the task of the Environment Spectator is to extract global semantic information of each snippet, we process the snippet through all convolutional blocks and several fully connected layers of the backbone network to obtain a single feature vector $f^{e}$, which captures information of all spatial dimensions of the snippet.
%\hl{Should move this to implementation details}
%For ActivityNet-1.3 dataset \cite{caba2015activitynet}, we employ C3D \cite{C3D} pre-trained on Kinetics-400 \cite{Kinetics} as our backbone network and remove its softmax and classification layers to keep a more semantic feature vector, which is in 2048 dimensions, $f^{e}\in\mathbb{R}^{2048}$.

%Contrarily, for the THUMOS-14 dataset \cite{THUMOS14}, we follow \cite{lin2018bsn, bmn, xu2020gtad} to employ TSN \cite{TSN2016ECCV} pre-trained on untrimmed action recognition track of ActivityNet-1.3 dataset. Additionally, to conduct a fair comparison with previous works, we keep the same settings as it was used in previous works to extract a 400-dimension feature vector from the last layer of TSN \cite{TSN2016ECCV}, $f^{e}\in\mathbb{R}^{400}$.

%\subsubsection{Actors Spectator}
\vspace{0.2cm}
\textbf{(ii) Actors Spectator}
\label{subsubsec:actors}
aims to semantically extract main actor(s) representation. An action cannot happen in the absence of a human (actor). However, when an action occurs, it does not necessarily signal that every actor in the scene has committed the action. First, the actors spectator detects all existing actors in the snippet by an \textit{actor localization} module. Then, an \textit{adaptive attention mechanism (AAM)} is proposed to adaptively select an arbitrary number of main actor(s) and extract their mutual relationships to represent them as a single feature vector.

\begin{figure}[t]
\centering
  \includegraphics[width=0.9\linewidth]{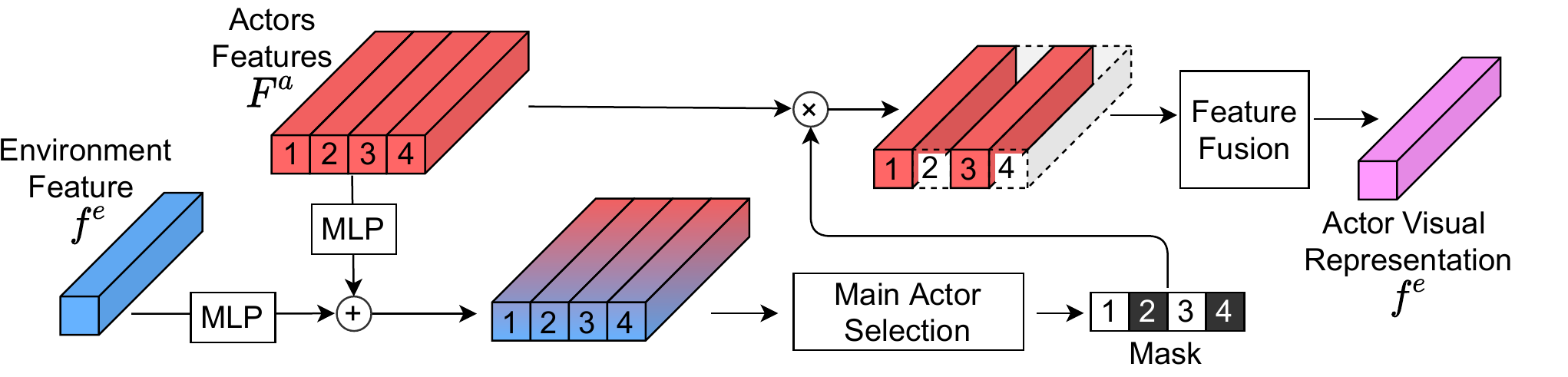}
  \vspace*{-0.1cm}
  \caption{Illustration of our proposed AAM. Given an environment feature $f^e$ and a set of actor features $F^a$, this module aims to select main actor features, followed by fusing arbitrary main actor features to obtain an actor visual representation $f^a$.}
  \vspace*{-0.4cm}
  \label{adaptive_attention}
\end{figure}

\noindent
\textit{\textbf{Actor Localization}}: 
To localize all actors in a $\delta$-frame snippet, we apply a human detector onto the middle frame of it with the assumption that, with a small $\delta$, the actors would not move fast enough to be mis-located. We denote  $\mathcal{B}=\{b_i\}_{i=1}^{N_B}$ as a set of detected human bounding boxes, where $N_B \geq 0$. Afterwards, each of the detected bounding boxes, $b_i$, is aligned onto feature map $\mathcal{M}$ (obtained by the 3D network backbone from environment spectator) using RoIAlign \cite{MaskRCNN_ICCV17} and then average-pooled into a single feature vector $f^a_i$. Finally, we obtain a set of actor features $F^a=\{f^a_i\}^{N_B}_{i=1}$. \\
\noindent
\textit{\textbf{Adaptive Attention Mechanism (AAM)}}:
Given $N_B$ detected actors, there are only a few of detected actors (called main actors) who actually contribute to the action if it presents. Because the number of main actors is unknown and continuously changes throughout the input video, we propose an adaptive attention mechanism (AAM) that inherits the merits from adaptive hard attention to select an arbitrary number of main actors and a soft self-attention mechanism \cite{attention_is_all_you_need} to extract relationships among them. AAM is described by pseudocode in Algorithm \ref{algo:aam} and illustrated in Fig. \ref{adaptive_attention}; more details are provided in the supplementary.

\begin{algorithm}[t]
\footnotesize
\DontPrintSemicolon
\SetNoFillComment
\KwData{Feature vector $f^e$ and features set $F^a$ represent environment and all actors that appear in input snippet, respectively.}
\KwResult{Feature vector $f^a$ represents main actors in input snippet.}
$\hat{f}^e \gets MLP_{\theta_e}(f^e)$ \tcp*{embed $f^e$ to common space with every $f^a_i$ in $F^a$}
set $S^a$ and $\tilde{F}^a$ to empty list \tcp*{$S^a$ will store scores of every actor}
set $\tilde{F}^a$ to empty list \tcp*{$F^a$ will store selected main actors}
\For{each $f^a_i$ in $F^a$}{
    $\hat{f}^a_i \gets MLP_{\theta_a}(f^a_i)$ \tcp*{embed $f^a_i$ to common space with $f^e$}
    $s^a_i \gets || \hat{f}^a_i \oplus \hat{f}^e ||_2$ to $S^a$ \tcp*{compute main actor score corresponding to $f^a_i$}
    append $s^a_i$ to $S^a$
}

$S^a \gets softmax(S^a)$ \tcp*{rescale scores to sum up to $1.0$}
$\tau \gets \frac{1}{|S^a|}$ \tcp*{compute adaptive threshold $\tau$}

\For{each $f^a_i$ in $F^a$}{
    \If{$S^a_i > \tau$}
    {
        append $f^a_i$ to $\tilde{F}^a$ \tcp*{select $f^a_i$ if its score higher than $\tau$}
    }
}
$f^a \gets self\_attention(\tilde{F}^a)$ \tcp*{fuse main actor features by self-attention \cite{attention_is_all_you_need}}
\caption{Adaptive Attention Mechanism (AAM) to extract representation of main actors in a snippet.}
\label{algo:aam}
\end{algorithm}

%Given a $\delta$-frame snippet, it displays that only a few actors among $N_B$ detected ones actually commit the action. Because the number of main actors is always unknown and may continuously change throughout the video duration, it is crucial to have an appropriate attention module to handle such an arbitrary number of actors. To address this issue, we propose AAM that inherits the merits from adaptive hard attention to select main actors and a soft self-attention mechanism to extract feature vectors from an arbitrary number of main actors.

%Because an arbitrary number of actors may simultaneously appear in a snippet, it is essential to attend only to main actors who are likely to commit the actions of interest for later action detection, instead of taking all actors into account. Therefore, adopting a hard attention module, which reasons on the actors and overall environment to select a number of actors which are possibly the main actors are very intuitively.

%However, because the number of main actors is always unknown and may continuously change throughout the video duration, it is crucial to have the hard attention module adaptively attend to an unfixed number of actors. The following are the steps to adopt the Adaptive Hard Attention Module \cite{adahard_eccv2018} into our model in order to attend to the main actors in every snippet.
\begin{figure*}[t]
\centering
  \includegraphics[width=0.8\linewidth]{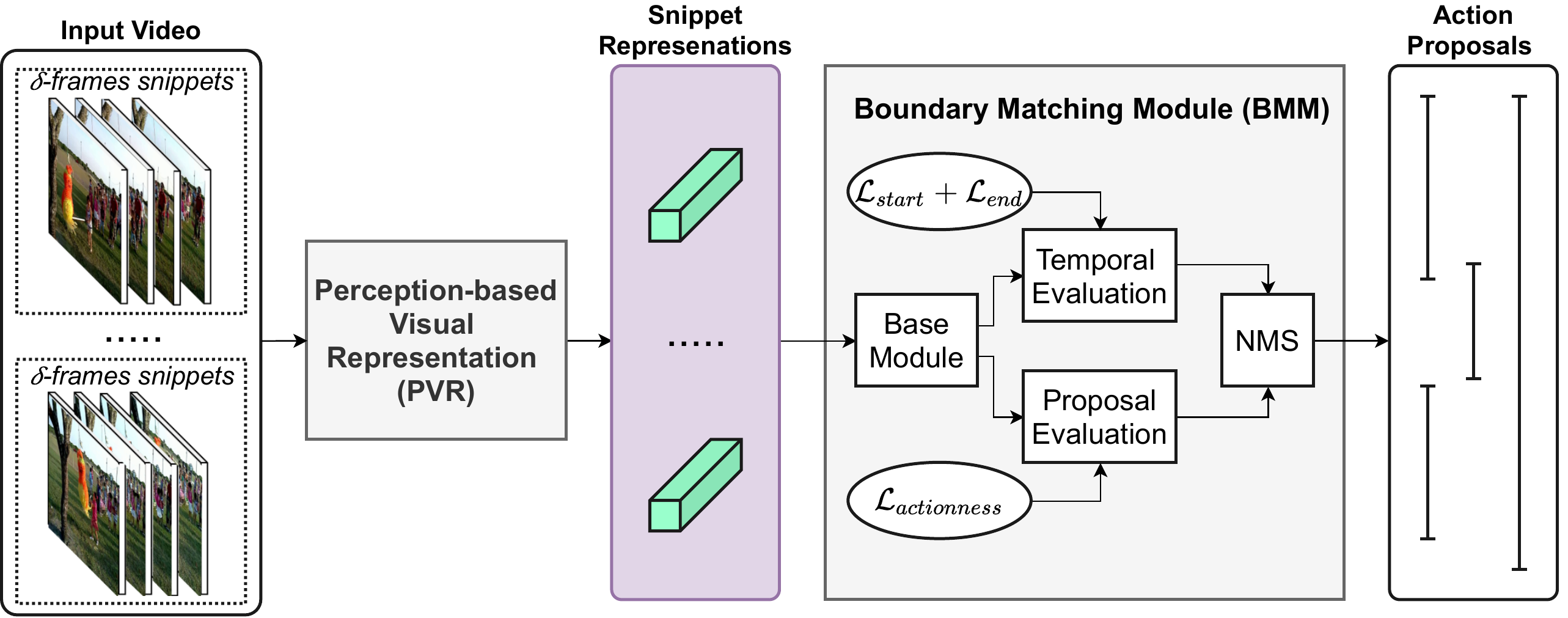}
  \vspace*{-0.1cm}
  \caption{The overall architecture of our proposed AEI, consisting of perception-based visual representation module (PVR), and boundary-matching module (BMM). }
  \vspace*{-0.4cm}
  \label{full_architecture}
\end{figure*}

\vspace{0.2cm}
\textbf{(iii) Actors-Environment Interaction Spectator}
This module aims to model the relations between environment feature $f^e$ and actors representation feature $f^a$, and then combine them into a single feature $f$. Herein, we employ the self-attention model \cite{attention_is_all_you_need} where $f^e$ and $f^a$ are the inputs. We denote $f_i$ as a visual representation for snippet $s_i$.

%After receiving environment feature $f^e$ from the Environment Spectator and actors representation feature $f^a$ from the Actors Spectator, the Actors-Environment Interaction Spectator is created to search for the relationship between both features and fuse them together into a single feature $f$ to represent the whole snippet. This agent adopts the same architecture of Transformer Encoder as in Actors Spectator for actors feature fusion, except that the input features set always has two elements i.e. $f^e$ and $f^a$.

%Striding through each snippet $s_i$ of the video, our \textcolor{red}{NAME} proceed through above three agents and produce a feature $f_i$ to represent the snippet $s_i$. $f_i$ captures both local information from main actors predicted by the hard attention module and the global information from the environment, both are fused with a proper ratio by the learned soft attention module in order to help the temporal action proposal module, which will be discussed below, achieve the SOTA performances.

\subsection{Boundary-Matching Module (BMM)}
\label{subsec:bmm}

BMM is responsible for generating action proposals, which are boundary-pairs of every possible action of interest appearing in the video. Our BMM contains three components: base module, temporal evaluation module, and proposal evaluation module as illustrated in Fig. \ref{full_architecture}. The base module aims to model the semantic relationship between snippets. The temporal evaluation module assesses each snippet $0\leq i \leq T$ in the video to estimate probabilities that any action starts or ends there, corresponding to $P^S_i$ and $P^E_i$, respectively. Meanwhile, the proposal evaluation module evaluates every interval $[i,j]$ in the video to estimate its actionness score, corresponding to $P^A_{i,d}$, where $d=j-i$.

At the inference stage, we search through $P^S$ and $P^E$ to select temporal locations $i$ whose $P^S_i$ or $P^E_i$ are local maximums to form sets of potential starting and ending temporal locations, respectively. Then, starting and ending locations $(s, e)$ from those lists that satisfy timing constraint (e.g. $s\leq e \leq T$) are paired and become a candidate proposal with a score $s=P^S_s \cdot P^E_e \cdot P^A_{s, e-s}$. 
Based on the timestamps and scores of candidate proposals, we finally apply NMS \cite{SoftNMS, NMS} to produce the final set of action proposals.

%any temporal location having a local maximum starting score (i.e. its starting score is higher than its nearest neighbors) is recorded to a possible starting points list. Similarly, an ending points list is formed by recording temporal locations having local maximum ending scores. Then, candidate proposals are generated by pairing starting and ending points that satisfy some upper and lower distancing thresholds. Finally, the score for a candidate proposal $p$ starting from $s_p$ and ending at $e_p$ is composed as follows:

% \begin{equation}
%     score_p = P_S[s_p] \cdot P_E[e_p] \cdot P_D[s_p, e_p]
% \end{equation}

% Afterward, a non-maximum suppression method \cite{SoftNMS, NMS} is applied to the set of candidate proposals to eliminate redundant proposals and produce the final set of action proposals.

In our paper, we have conducted BMM under two different network architectures: CNN-based and GCN-based. Our CNN-based BMM, called \textbf{AEI-B}, is leveraged by \cite{bmn} where the base module is comprised of 1D convolutional layers to learn and extract the temporal relations between snippets. On the other hand, our GCN-based BMM, called \textbf{AEI-G}, is leveraged by \cite{xu2020gtad} to extract not only local relations, but also the relations of snippets that share close semantic features.

\subsection{Training Methodology}
\label{subsec:train_method}

Given a list of $N_G$ ground-truth action segments $G=\{g_i=(g^s_i, g^e_i)\}_{i=1}^{N_G}$ of input video $\mathcal{V}$, we generate the ground-truth starting labels $L^S \in [0,1]^T$, ending labels $L^E \in [0,1]^T$ and actionness labels $L^A \in [0,1]^{T\times D}$ ($D$ is a pre-defined maximum proposal length). $L^S_i$ (or $L^E_i$) carrying a value of $1$ means that there is a ground-truth starting (or ending) boundary of any action at $i$ and vice versa. Likewise, $L^A_{i,d}$ carrying a value of $1$ means that there is a ground-truth action starts at $i$ with a length of $d$.
To train our AEI network with the ground-truth labels, we define the loss function $\mathcal{L}_{AEI}$ as follows:
%is comprised of three loss functions: $\mathcal{L}_{start}$,  $\mathcal{L}_{end}$, and $\mathcal{L}_{actionness}$. The first two losses aim to localize starting frame and ending frame, while the last one generates the actionness scores matrix. Mathematically, 
%We follow \cite{lin2018bsn, bmn, xu2020gtad} to define our loss function to train the TAPG pipeline. The total loss function is comprised of three loss functions e.g. $\mathcal{L}_{start}$, $\mathcal{L}_{end}$ and $\mathcal{L}_{actionness}$. The two formers are designated to train the pipeline on classifying precise starting temporal points, ending temporal points and the latter is designated to train the pipeline to generate accurate the actionness scores matrix, respectively. The following are equations defining the loss functions and total loss function:
\begin{equation*}
    \mathcal{L}_{AEI} = \mathcal{L}_{start}(P^S, L^S) + \mathcal{L}_{end}(P^E,L^E) + \mathcal{L}_{actionness}(P^A,L^A)
\end{equation*}
%\begin{equation*}
%\begin{aligned}[c]
%\mathcal{L}_{start} & = \mathcal{L}_{wb}(P^S, L^S) \\ \mathcal{L}_{end} & = \mathcal{L}_{wb}(P^E, L^E)
%\end{aligned}
%\qquad\qquad\qquad
%%\qquad\Longleftrightarrow\qquad
%\begin{aligned}[c]
%\mathcal{L}_{actionness} & = \mathcal{L}_{wb}(P^D, L^D) + \lambda_{reg} \mathcal{L}_2(P^D, L^D) \\
%\end{aligned}
%\end{equation*}
% \begin{equation}
% \begin{split}
%     \mathcal{L}_{start} & = \mathcal{L}_{wb}(P_S, L_S) \text{         }\mathcal{L}_{end} = \mathcal{L}_{wb}(P_E, L_E) \\
%     \mathcal{L}_{actionness}  = \mathcal{L}_{wb}(P_D, L_D) + \lambda_{reg} \mathcal{L}_2(P_D, L_D)\text{         }
%     \mathcal{L}_{AEI} & = \mathcal{L}_{start} + \mathcal{L}_{end} + \mathcal{L}_{actionness}
% \end{split}
% \end{equation}
We use weighted binary log-likelihood loss $\mathcal{L}_{wb}$ for $\mathcal{L}_{start}$ and $\mathcal{L}_{end}$, which is defined as follows:
\begin{equation}
\small
    \mathcal{L}_{wb}(P, L) =  \sum^{N}_{i=1}\left[\frac{L_i}{N^+}\log P_i + \frac{(1-L_i)}{N^-}\log (1-P_i)\right]
\label{eq:L_wb}
\end{equation}

where $N^+$ and $N^-$ are the number of positives and negatives in ground-truth labels, respectively. Conversely, $\mathcal{L}_{actionness}(P,L)=\mathcal{L}_{wb}(P,L)+\lambda \mathcal{L}_2(P,L)$, where $\mathcal{L}_2$ is the mean squared error loss and $\lambda$ is set to $10$.

To reduce time cost in training phase of our proposed AEI network, actors features set $F^a$ and environment feature $f^e$ for actors spectator and environment spectator, respectively, are extracted in advance. Then, the AAM and Interaction Spectator of PVR module is trained with BMM module in an end-to-end fashion.

\section{Experiments}
%We conduct the experiments and comparisons between our AEI and SOTA methods on both TAPG and TAD tasks.
% Our BMM (the second module in AEI) is built upon two network architectures: CNN-based and GCN-based. We denote these architectures as \textbf{AEI-B} and \textbf{AEI-G}, respectively.
% \newline
% \newline
\noindent
\textbf{Datasets}

\noindent
We conduct experiments on ActivityNet-1.3 \cite{caba2015activitynet} and THUMOS-14 \cite{THUMOS14} datasets. The former contains 20,000 videos with 200 annotated activities while the latter consists of 414 videos with 20 actions. For both datasets, we follow previous works \cite{lin2018bsn,bmn, dbg} to preprocess videos with the snippet length set to $\delta=16$.

\begin{table}[t]
\centering
\caption{\textbf{TAPG} comparisons in terms of AR@AN and AUC between our AEI and other SOTA methods on \textbf{ActivityNet-1.3}.}
\vspace*{0.2cm}
\resizebox{\linewidth}{!}{
\begin{tabular}{c||ccccccccc|cc}
\centering
\multirow{2}{*}{Metrics} & BSN & MGG & MR & BMN & DBG & BSN++ & TSI++ & AEN & ABN & \multirow{2}{*}{AEI-B} & \multirow{2}{*}{AEI-G} \\
& \cite{lin2018bsn} & \cite{liu2019multi} & \cite{MR_eccv2020} & \cite{bmn} & \cite{dbg} & \cite{BSN++} & \cite{tsi_accv} & \cite{KhoaVo_ICASSP} & \cite{KhoaVo_Access} & &  \\ \hline \hline
AR@100 (val)& 74.16 & 74.54 & 75.27 & 75.01 & 76.65 & 76.52 & 76.31 & 75.65 & 76.72 & \textbf{77.25} & \underline{\textit{77.24}} \\
AUC (val)   & 66.17 & 66.43 & 66.51 & 67.10 & 68.23 & 68.26 & 68.35 & 68.15 & 69.16 & \underline{\textit{69.43}} & \textbf{69.47} \\
AUC (test)  & 66.26 & 66.47 & - & 67.19 & 68.57 & - & 68.85 & 68.99 & 69.26 & \underline{\textit{69.94}} & \textbf{70.09} \\
\bottomrule
\end{tabular}
}
\label{activitynet_proposal1}
\end{table}

\noindent
\textbf{Metrics}
\newline 
In TAPG, AR@AN and AUC are popular metrics to benchmark the performance of competing methods. The former is the average recall (AR) computed with the average number of proposals (AN) kept by each video. The latter is the score of the area under AR versus the AN curve. In ActivityNet-1.3, AR@100 and AUC are mainly used. On the other hand, in THUMOS-14, only AR@AN is used to compare between methods; however, multiple AN is selected from a list of [50, 100, 200, 500, 1000].
\newline\indent
In TAD, both ActivityNet-1.3 and THUMOS-14 use mean Average Precision (mAP) to benchmark methods in this problem. ActivityNet-1.3 uses tIoU thresholds of \{0.5, 0.75, 0.95\} and average mAP, while THUMOS-14 uses tIoU thresholds of \{0.3, 0.4, 0.5, 0.6, 0.7\} for evaluation.

\noindent
\textbf{Implementation Details}

\noindent
For all experiments on both ActivityNet-1.3 \cite{caba2015activitynet} and THUMOS-14 \cite{THUMOS14}, we employ C3D \cite{C3D} network pre-trained on Kinetics-400 \cite{Kinetics} as the backbone network to extract features from video snippets. The features extracted from C3D backbone have 2048 dimensions.
\newline \indent
In the actors spectator, for actor localization, we employ a Faster-RCNN model \cite{FasterRCNN} pre-trained on COCO \cite{cocodataset} dataset to detect humans as discussed in \ref{subsubsec:actors}. To train our AEI network, we utilize Adam optimizer with the initial learning rate set to 0.0001 and 0.001 for ActivityNet-1.3 and THUMOS-14, respectively. 
%Empirically, We find that these settings for training phase give the best performances for our AEI network.
\newline \indent
In TAPG, Soft-NMS \cite{SoftNMS} is applied in post-processing for all experiments on ActivityNet-1.3, while on THUMOS-14, both Soft-NMS \cite{SoftNMS} and NMS \cite{NMS} are evaluated. In TAD, following \cite{lin2018bsn, bmn}, we use NMS \cite{NMS} for post-processing on both datasets.
\newline \indent
In the following experiments, we highlight the best performance in \textbf{bold} and the second-best performance in \underline{\textit{italic}}.

%\noindent
%\textbf{Temporal Action Proposal Generation (TAPG)}
\subsection{Temporal Action Proposal Generation (TAPG)}

%In this section, we conduct a performance comparison between our AEI and other SOTA methods on the TAPG task. 
Table \ref{activitynet_proposal1} demonstrates the comparison on ActivityNet-1.3 validation and testing sets. Based on the results, it can be observed that the performances of our methods, \text{AEI-B} and \text{AEI-G}, are standing out against those of SOTA methods in terms of AR@100 and AUC by large margins. Table \ref{thumos_proposal} presents the comparison of SOTA TAPG methods on THUMOS-14 dataset. Compared to SOTA approaches, our AEI obtains better performance on all AR@ANs metrics regardless of the architecture of BMM. From Table \ref{activitynet_proposal1}, \ref{thumos_proposal}, we empirically observe that AEI-G, which employs GCN to model the relationship between snippets, marginally surpasses  AEI-B on TAPG.

%\begin{table}[t]
%\centering
%\caption{\textbf{TAD} comparisons on \textbf{ActivityNet-1.3} in terms of mAP@tIoU and mAP, where the proposals are combined with video-level classification results generated by \cite{action_protocol}.}
%\vspace*{0.2cm}
%\resizebox{\linewidth}{!}{
%\begin{tabular}{c||cccccccccc|cc}
%\multirow{2}{*}{Metrics} & CDC & SSN & BSN & BMN & GTAD & P-GCN & MR & ABN & TSI++ & GTAN & \multirow{2}{*}{AEI-B} & \multirow{2}{*}{AEI-G} \\
%& \cite{CDC} & \cite{SSN} & \cite{lin2018bsn} & \cite{bmn} & \cite{xu2020gtad} & \cite{pgcn_cvpr2020} & \cite{MR_eccv2020} & \cite{KhoaVo_Access} & \cite{tsi_accv} & \cite{gtan_cvpr2019} &  &  \\
%\hline \hline
%0.5 & 43.8 & 39.1 & 46.5 & 50.1 & 50.4 & 42.9 & 43.5 & 51.8 & 51.2 & \textbf{52.6} & 52.3 & \underline{\textit{52.4}}\\
%0.75 & 25.9 & 23.5 & 30.0 & \underline{\textit{34.8}} & 34.6 & 28.1 & 33.9 & 34.2 & \textbf{35.0} & 34.1 & 34.5 & 34.5 \\
%0.95 & 0.2 & 5.5 & 8.0 & 8.3 & 9.0 & 2.5 & 9.2 & \textbf{10.3} & 6.6 & 8.9 & \underline{\textit{9.7}} & 9.6 \\
%Average & 22.8 & 24.0 & 30.0 & 33.9 & 34.1 & 27.0 & 30.1 & 34.2 & 34.2 & \underline{\textit{34.3}} & \textbf{34.7} & \textbf{34.7} \\
%\bottomrule
%\end{tabular}
%}
%\label{activitynet_detection}
%\end{table}

\begin{table}[t]
\centering
\caption{\textbf{TAD} comparisons on \textbf{ActivityNet-1.3} in terms of mAP@tIoU and mAP, where the proposals are combined with video-level classification results generated by \cite{action_protocol}.}
\vspace*{0.2cm}
\resizebox{\linewidth}{!}{
\begin{tabular}{c||cccccccc|cc}
\multirow{2}{*}{Metrics} & BSN & BMN & GTAD & P-GCN & MR & ABN & TSI++ & GTAN & \multirow{2}{*}{AEI-B} & \multirow{2}{*}{AEI-G} \\
& \cite{lin2018bsn} & \cite{bmn} & \cite{xu2020gtad} & \cite{pgcn_cvpr2020} & \cite{MR_eccv2020} & \cite{KhoaVo_Access} & \cite{tsi_accv} & \cite{gtan_cvpr2019} &  &  \\
\hline \hline
0.5 & 46.5 & 50.1 & 50.4 & 42.9 & 43.5 & 51.8 & 51.2 & \textbf{52.6} & 52.3 & \underline{\textit{52.4}}\\
0.75 & 30.0 & \underline{\textit{34.8}} & 34.6 & 28.1 & 33.9 & 34.2 & \textbf{35.0} & 34.1 & 34.5 & 34.5 \\
0.95 & 8.0 & 8.3 & 9.0 & 2.5 & 9.2 & \textbf{10.3} & 6.6 & 8.9 & \underline{\textit{9.7}} & 9.6 \\
Average & 30.0 & 33.9 & 34.1 & 27.0 & 30.1 & 34.2 & 34.2 & \underline{\textit{34.3}} & \textbf{34.7} & \textbf{34.7} \\
\bottomrule
\end{tabular}
}
\label{activitynet_detection}
\end{table}

%\begin{table}[thb]
%\centering
%\caption{\textbf{TAD} performance comparisons on \textbf{ActivityNet-1.3} in terms of mAP@tIoU and average mAP, where the proposals are combined with video-level classification results generated by \cite{action_protocol}. The best and the second-best performances are shown in \textbf{bold} and \underline{\textit{italic}}, respectively.}
%\begin{tabular}{l l l l l l}
%Method                      & 0.5   & 0.75  & 0.95 & Average \\
%\hline \hline
%CDC \cite{CDC}              & 43.8 & 25.9 & 0.2 & 22.8 \\
%SSN \cite{SSN}              & 39.1 & 23.5 & 5.5 & 24.0 \\
%BSN \cite{lin2018bsn}       & 46.5 & 30.0 & 8.0 & 30.0 \\
%BMN \cite{bmn}              & 50.1 & \underline{\textit{34.8}} & 8.3 & 33.9 \\
%GTAD \cite{xu2020gtad}      & 50.4 & 34.6 & 9.0 & 34.1 \\ 
%P-GCN\cite{pgcn_cvpr2020}   & 42.9 & 28.1 & 2.5 & 27.0 \\
%MR\cite{MR_eccv2020}        & 43.5 & 33.9 & 9.2 & 30.1 \\
%TSI++~\cite{tsi_accv}       & 51.2 & \textbf{35.0} & 6.6 & 34.2 \\
%GTAN~\cite{gtan_cvpr2019}   & \textbf{52.6} & 34.1 & 8.9 & 34.3 \\
%\hline 
%% \textbf{S-AEI-B}  & 51.78 & 34.18 & 9.29 & 34.22 \\
%% \textbf{S-AEI-G}  & 51.93 & 34.05 & 9.36 & 34.26 \\
%\textbf{AEI-B}  & 52.3 & 34.5 & \textbf{9.7 }& \textbf{34.7} \\
%\textbf{AEI-G}  & \underline{\textit{52.4}} & 34.5 & \underline{\textit{9.6}} & \textbf{34.7} \\
%\bottomrule
%\end{tabular}
%\label{activitynet_detection}
%\end{table}

\begin{figure}[t]
  \centering
  \includegraphics[width=0.8\textwidth]{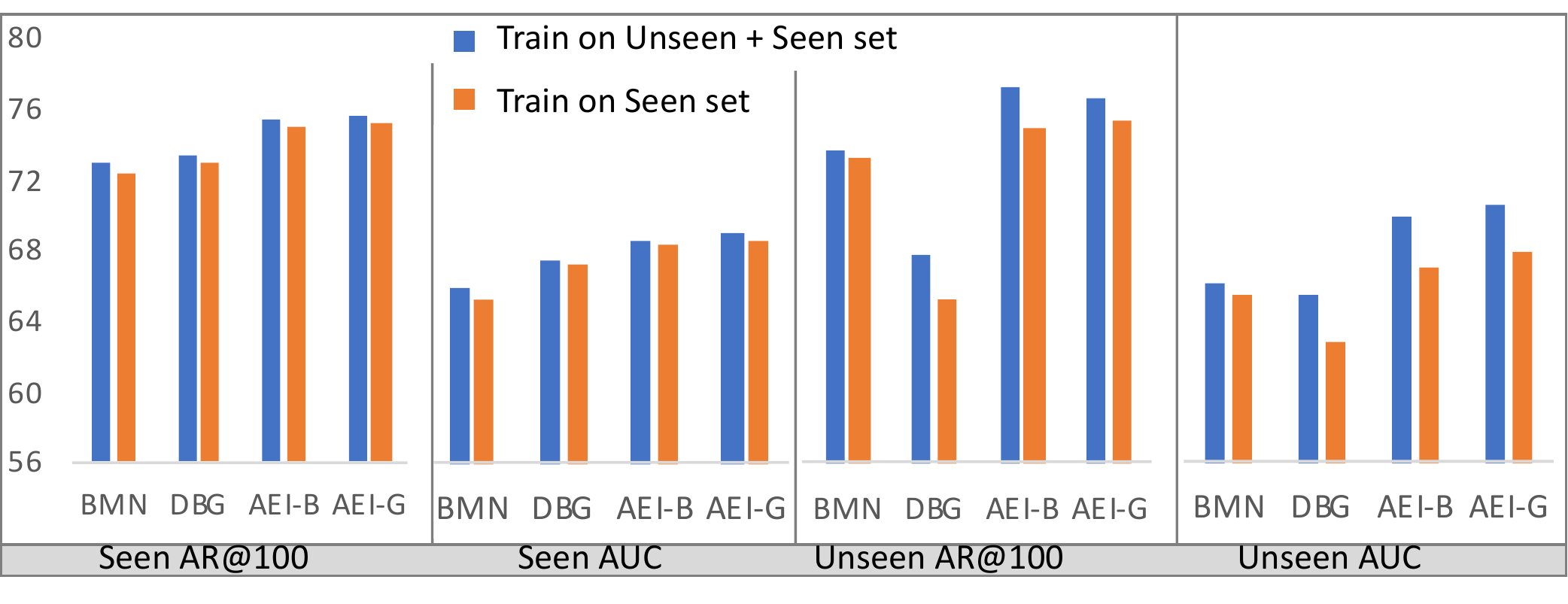}
  \vspace*{-0.1cm}
  \caption{\textbf{Generalizability} evaluation and comparisons with BMN \cite{bmn} and DBG \cite{dbg} on ActivityNet 1.3 in terms of AR@100 and AUC. Methods are trained on \textit{Unseen+Seen} (blue columns) and \textit{Seen} training sets (orange columns), respectively; and are evaluated on \textit{Seen} (first two charts) and \textit{Unseen} (last two charts) validation sets.}
  \label{fig:Generalizability}
  \vspace*{-0.4cm}
\end{figure}

Generalizability is also considered as an important criterion to evaluate a method in TAPG. 
%In this study, which is held on ActivityNet-1.3 \cite{caba2015activitynet}, videos belonging to two non-overlapped action class groups of "Sports, Exercises, and Recreation" and "Socializing, Relaxing, and Leisure" are collected into \textit{Seen} and \textit{Unseen} subsets, respectively.
Following the same experiment setup in \cite{bmn, dbg}, our AEI is trained on \textit{Unseen+Seen} and \textit{Seen} training sets, separately, and then is evaluated on the \textit{Seen} and \textit{Unseen} validation sets, separately as illustrated in Fig.~\ref{fig:Generalizability}. %Fig.~\ref{fig:Generalizability} shows the generalizability comparison between AEI and the SOTA methods (BMN\cite{bmn}, DBG\cite{dbg}) in terms of AR@AN and AUC. 
The performances on \textit{Seen} validation set is shown in the first two charts, whereas the performances on \textit{Unseen} validation set is given in the last two charts.
Fig. \ref{fig:Generalizability} shows that our AEI achieves good performances on \textit{Seen} validation set with an acceptable drop on \textit{Unseen} validation set on both training configurations, suggesting that our AEI is highly generalizable to unseen action types.
%Our AEI not only achieves good performance on \textit{Seen} validation set when trained by \textit{Unseen+Seen} and \textit{Seen}, but also obtains remarkable scores on \textit{Unseen} validation set regardless of BMM network architectures. This implies that our AEI has a good generalization and can even localize action classes that it never sees in training phase.

%\begin{figure}
%  \begin{minipage}[c]{0.37\textwidth}
%   \caption{\textbf{Generalizability} evaluation on ActivityNet 1.3 with AR@100 and AUC. AEI is trained on \textit{Unseen+Seen} training set and \textit{Seen} training set. AEI is tested on \textit{Seen} (first two charts) and \textit{Unseen} (last two charts) validation sets.}
%  \label{fig:Generalizability}
%  \end{minipage}\hfill
%  \begin{minipage}[c]{0.60\textwidth}
%    \includegraphics[width=\textwidth]{Figures/generalizability.pdf}
%  \end{minipage}
%\end{figure}

\subsection{Temporal Action Detection (TAD)}

Following the experiment settings in \cite{lin2018bsn, bmn, xu2020gtad}, we adopt top-1 video-level classification results generated by the method in \cite{action_protocol} on ActivityNet-1.3 to label the proposals produced by our method. We use top-2 video-level classification results generated by UntrimmedNet \cite{untrimmetNet} to label proposals generated by our method on THUMOS-14.

Table \ref{activitynet_detection} illustrates TAD performance comparison between AEI and other SOTA methods on ActivityNet-1.3 validation set. The results emphasize that our methods outperform SOTA methods in spite of CNN-based BMM or GCN-based BMM. The experiment results on THUMOS-14 test set in Table \ref{thumos_detection} demonstrate that our AEI-B and AEI-G are superior to other SOTA methods on most of the metrics regardless of UntrimmedNet \cite{untrimmetNet} or P-GCN \cite{pgcn_cvpr2020} classifiers. From Table \ref{activitynet_detection} and \ref{thumos_detection}, we empirically notice that both AEI-B and AEI-G obtain comparable TAD performance.

\begin{table}[tb]
    \begin{minipage}[t]{.55\linewidth}
        \centering
        \caption{\textbf{TAPG} comparisons on \textbf{THUMOS-14} in terms of AR@AN, where SNMS represents Soft-NMS \cite{SoftNMS}.}\vspace{0.2cm}
        \resizebox{\linewidth}{!}{
        \begin{tabular}{c c c c c c }
        Method & @50 & @100 & @200 & @500 & @1000 \\
        \hline \hline
        T-TAG \cite{anchor_3}    & 18.55 & 29.00 & 39.61 & -     & -     \\
        CTAP  \cite{CTAP}        & 32.49 & 42.61 & 51.97 & -     & -     \\
        BSN \cite{lin2018bsn}    & 37.46 & 46.06 & 53.21 & 60.64 & 64.52 \\
        MGG  \cite{liu2019multi} & 39.93 & 47.75 & 54.65 & 61.36 & 64.06 \\
        BMN \cite{bmn}           & 39.36 & 47.72 & 54.70 & 62.07 & 65.49 \\
        DBG+NMS\cite{dbg}           & 40.89 & 49.24 & 55.76 & 61.43 & 61.95 \\
        DBG+SNMS \cite{dbg}      & 37.32 & 46.67 & 54.50 & 62.21 & 66.40 \\ 
        TSI++\cite{tsi_accv}     & 42.30 & 50.51 & 57.24 & 63.43 & -     \\
        MR\cite{MR_eccv2020}     & 44.23 & 50.67 & 55.74 & -     & -     \\
        ABN+NMS\cite{KhoaVo_Access}  & 44.89 & 51.86 & 57.36 & 61.67 & 62.59 \\
        ABN+SNMS\cite{KhoaVo_Access}  & 40.87 & 49.09 & 56.24 & 63.53 & 67.29 \\
        \hline
        \textbf{AEI-B+NMS} & \textbf{45.74} & \underline{\textit{52.39}} & 57.74 & 62.49 & 63.38 \\
        \textbf{AEI-G+NMS} & 45.12 & \textbf{52.81} & \underline{\textit{57.94}} & 62.11 & 63.17 \\ 
        \textbf{AEI-B+SNMS} & 44.97 & 50.13 & 57.34 & \underline{\textit{64.43}} & \underline{\textit{67.78}} \\
        \textbf{AEI-G+SNMS} & \underline{\textit{45.31}} & 51.12 & \textbf{58.19} & \textbf{64.58} & \textbf{67.96} \\
        \bottomrule
        \end{tabular}
        }
        \label{thumos_proposal}
    \end{minipage}%
    \hspace{0.02\linewidth}
    \begin{minipage}[t]{.44\linewidth}
        \centering
        \caption{\textbf{TAD} comparisons on \textbf{THUMOS-14} in term of mAP@tIoU. $^*$ indicates P-GCN classifier ($2^{nd}$ group); otherwise, UntrimmedNet classifier.}\vspace{0.2cm}
        %The first group uses UntrimmedNet as a classifier whereas the second group use G-PCN as a classifier}
        \resizebox{\linewidth}{!}{
        \begin{tabular}{l l l l l l l} 
        Method  & 0.7   & 0.6  & 0.5   & 0.4   & 0.3 \\ \hline
        \hline
        % SST \cite{SST_CVPR2017}    & 4.7  & 10.9 & 20.0 & 31.5 & 41.2 \\
        T-TAP\cite{SST_CVPR2017}   & 6.3  & 14.1 & 24.5 & 35.3 & 46.3 \\
        BSN \cite{lin2018bsn}      & 20.0 & 28.4 & 36.9 & 45.0 & 53.5 \\
        BMN \cite{bmn}             & 20.5 & 29.7 & 38.8 & 47.4 & 56.0 \\
        MGG \cite{liu2019multi}    & 21.3 & 29.5 & 37.4 & 46.8 & 53.9 \\
        DBG \cite{dbg}             & 21.7 & 30.2 & 39.8 & 49.4 & 57.8 \\
        GTAD \cite{xu2020gtad}     & \textbf{23.4} & 30.8 & 40.2 & 47.6 & 54.5 \\
        TSI++\cite{tsi_accv}       & 22.4 & 33.2 & 42.6 & \underline{\textit{52.1}} & \textbf{61.0} \\
        GTAN~\cite{gtan_cvpr2019}  & -- & -- & 38.8 & 47.2 & 57.8 \\
        %ABN\cite{KhoaVo_Access}  & 25.6 & 37.0 & 46.1 & 54.0 & 59.9 \\
        \hline
        
        %\textbf{S-AEI-B} & 23.3 & 34.6 & 43.8 & 51.8 & 57.8 \\% threshold 0.45, 3000 proposals -2
        %\textbf{S-AEI-G} & & & & & \\% threshold 0.45, 3000 proposals
        \textbf{AEI-B} & \textbf{23.4} & \textbf{35.9} & \textbf{44.7} & \textbf{52.7} & \underline{\textit{58.7}} \\% threshold 0.45, 5000 proposals
        \textbf{AEI-G} & \underline{\textit{22.9}} & \underline{\textit{34.2}} & \underline{\textit{43.4}} & 51.6 & 57.4 \\% threshold 0.45, 3000 proposals
        \hline \hline
        BSN$^*$\cite{lin2018bsn}   & --   & --   & 49.1 & 57.8 & 63.6 \\
        GTAD$^*$ \cite{xu2020gtad} & \underline{\textit{22.9}} & 37.6 & 51.6 & \underline{\textit{60.4}} & 66.4 \\
        MR$^*$\cite{MR_eccv2020}     & \textbf{28.5} & \textbf{38.0} & 45.4 & 50.7 & 53.9 \\
        \hline
        % \textbf{S-AEI-B}$^*$ & 23.4 & 36.5 & 50.6 & 60.4 & 66.4 \\
        % \textbf{S-AEI-G}$^*$ & 23.3 & 37.2 & 51.0 & 60.2 & 66.7 \\
        \textbf{AEI-B}$^*$ & 22.4 & 37.8 & \textbf{52.1} & \textbf{60.6} & \underline{\textit{67.3}} \\
        \textbf{AEI-G}$^*$ & 22.3 & \underline{\textit{37.9}} & \underline{\textit{52.0}} & \underline{\textit{60.4}} & \textbf{67.6} \\
        \bottomrule
        \end{tabular}
        }
        \label{thumos_detection}
    \end{minipage} 
\end{table}

%\noindent
%\textbf{Ablation Study}
\subsection{Ablation Study}

\begin{figure}[t]
\centering
  \includegraphics[width=0.75\linewidth]{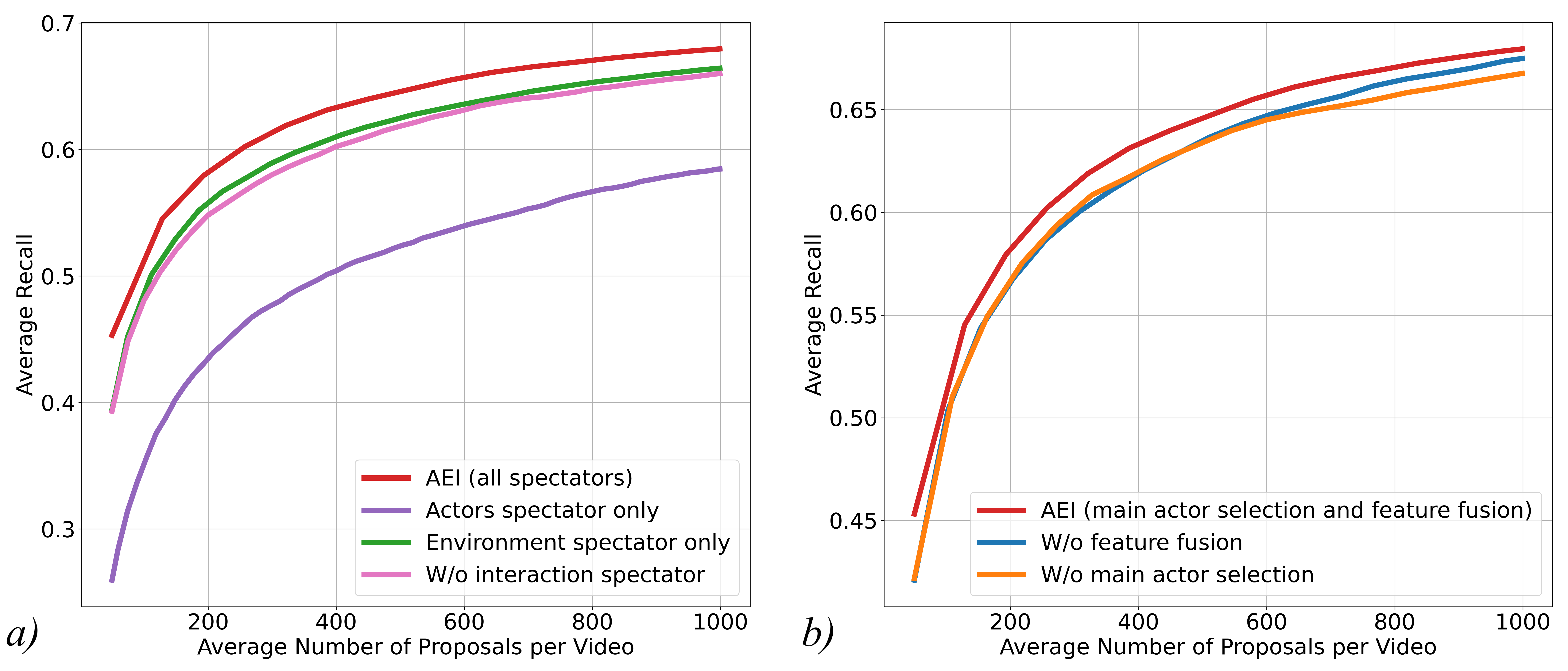}
  \vspace*{-0.1cm}
  \caption{TAPG comparisons on different AEI configurations: (a) either only environment or only actors spectator or both; (b) either only main actor selection or only feature fusion or both.}.
  %\caption{TAPG comparisons on different AEI configurations: (a) environment spectator vs. actors spectator vs. AEI (both environment and actor); (b) main actor selection (w/o feature fusion) vs. feature fusion (w/o main actor selection) vs. AEI (main actor selection and feature fusion)}.
\label{ablation_fig}
  \vspace*{-0.4cm}
\end{figure}

We further conduct a detailed ablation study on THUMOS-14 dataset to evaluate the contributions of different components of the proposed AEI framework. We conduct two ablation studies as shown in Fig.~\ref{ablation_fig} on TAPG and THUMOS-14 in terms of AR@ANs.

First, we evaluate the contribution of each spectator to the overall performance of our proposed PVR (described in Section~\ref{subsec:pvr}), i.e., environment spectator, actors spectator, and interaction spectator. As illustrated in Fig.~\ref{ablation_fig} (a), "Environment spectator only", which only focuses on global information, plays an important role in TAPG, whereas "Actors spectator only", which takes only local information of the main actor(s) into account, achieves adequate performance. "W/o interaction spectator", which withdraws the interaction spectator by simply fusing global and local information using an averaging operation, gives an undesired performance that is even lower than "Environment only". The complete proposed model, e.g., "AEI (all spectators)", gives the best result thanks to the interaction spectator adaptively fusing global feature from environment spectator and local feature from actors spectator.

In addition, we also evaluate the effectiveness of main actor selection and feature fusion in our proposed AAM. Fig.~\ref{ablation_fig} (b) shows the performance of the network without each of these components. In the "AEI w/o feature fusion" settings, we use an average pooling layer to fuse features obtained from main actor selection component. As illustrated, both configurations achieve similar performance with AN below 600, while main actor selection component plays a slightly more significant role than feature fusion component. This implies that having an appropriate main actor selection contributes significantly to the entire network.

\section*{Conclusion}
In this paper, we proposed a novel actors-environment interaction (AEI) network to simulate human perceiving ability in the temporal action proposals generation. Our AEI contains two modules: perception-based visual representation (PVR) and boundary-matching (BMM). PVR aims to extract visual representation of each snippet. To achieve the human perceiving ability, PVR is equipped with three spectators, each of which learns to perceive input snippet at a unique aspect, e.g. environment, main actors, and actors-environment interactions. An adaptive attention mechanism (AAM) is proposed in actors spectator to select an arbitrary number of main actor(s) in the snippet as well as learning the relationships between them.

Extensive experiments are conducted on ActivityNet-1.3 and THUMOS-14 datasets on TAPG and TAD tasks, which demonstrate that our proposed AEI outperforms SOTA methods regardless of BMM architecture (e.g., CNN-based or GCN-based). These results prove that replicating human perceiving ability in video understanding is a promising track to follow and further explore in the future. 

Beside three proposed spectators in PVR, we can include additional spectators to observe human body parts and the interaction between them with objects to better handle egocentric videos, in which the main actor who perform the action does explicitly appear.

\newpage

\section*{Acknowledgment}
This material is based upon work supported by the National Science Foundation under Award No. OIA-1946391; partially funded by Gia Lam Urban Development and Investment Company Limited, Vingroup and supported by Vingroup Innovation Foundation (VINIF) under project code VINIF.2019.DA19.

\bibliography{egbib}
\end{document}